# A New Method for Removing the Moire' Pattern from Images

Seyede Mahya Hazavei
Hamedan University of Technology
Hamedan, Iran
m.hazavei@yahoo.com

Hamid Reza Shahdoosti
Hamedan University of Technology
Hamedan, Iran
h.doosti@hut.ac.ir

*Abstract*— **During the last decades, denoising methods have attracted much attention of researchers. The conventional method for removing the Moire' pattern from images is using notch filters in the Frequency-domain. In this paper a new method is proposed that can achieve a better performance in comparison with the traditional method. Median filter is used in some part of spectrum of noisy images to reduce the noise. At the second part of this paper, to demonstrate the robustness of the proposed method, it is implemented for some noisy images that have moire' pattern. Experiments on noisy images with different characteristics show that the proposed method increases the PSNR values compared with previous methods.**

## I. INTRODUCTION

Denoising techniques are the most important parts of signal processing, that can be achieved by different ways. There are several kinds of noise, namely, salt and pepper noise, speckle noise, Gaussian white noise, Localvar noise, etc. Linear filtering is one of the most widely used frameworks for many image processing tasks. Apart from the desired noise reduction and scale selection, this technique has two major drawbacks: 1.details and edges are smoothed, and 2. on account of the absence of a natural basis for vector ordering, extension to multi-channel data is done by applying the operation on the channels separately. Hence the correlation between the channels is neglected in this method. To tackle these problems, several nonlinear filters have been proposed in literature [1]. In the following, some of these nonlinear filters are reviewed.

### A. Mode filtering

The distribution of the pixel values in a neighborhood, i.e. a local histogram, contains significant information about the zero order local image structure [1]. All maxima in a distribution are called modes. The global mode is the highest maximum of the distribution and hence is the most occurring value in the neighborhood. Local mode is the mode found by incorporating the pixel value in the center of the neighborhood as a-priori knowledge to the iteration procedure. The local mode has the property of noise reduction without the two drawbacks mentioned before.

### B. Median filters

The median filter is a nonlinear filtering technique which is widely used in the context of image denoising. This is due to the fact that under certain conditions, it preserves edges while removing noise. The main idea of the 2D median filter is to run through the image pixel by pixel, and replace the value of each pixel with the median of neighboring pixels. The neighbors' pattern is called the "window", which slides pixel by pixel, over whole of the image. For 1D (one dimensional) signals, the most obvious window is just the first few following and preceding samples, whereas for 2D (or higher-dimensional) signals such as images, one can use more complex window patterns such as "cross" or " box" patterns. Note that if the window is composed of an odd number of entries, then the median filter can be defined easily: it obtains the middle value of all the entries in the window which are numerically sorted. Median filtering is a useful tool for reducing salt and pepper noise in an noisy image [2], [3].

### C. Total variation algorithm

Total variation (TV) model was introduced by Rudin, Osher, and Fatemi in 1992 [4]. The total variation of a signal measures the amount of signal changes between signal values. That is, the integral of the absolute gradient of the signal. If $\Omega$ be a bounded region in $R^2$, the total variation of a function $u \in \Omega$ is defined as [5]:

$$TV(u) = \int_\Omega \|\nabla u\| dA \qquad (1)$$

where $\nabla u = (\partial u / \partial x_1 + \partial u / \partial x_2)$.

The goal of total variation denoising is to find an approximation, that has smaller total variation than the image but is "close" to the noisy signal. For the purpose of denoising, a conventional TV technique is based on minimizing the following function:

$$F(u) = \int_\Omega \|\nabla u\| dA + \frac{1}{2} \int_\Omega (u - u_0)^2 dA \qquad (2)$$

where u is the estimated image and $u_0$ is the noisy image. Eq. (2) can be solved with different approaches [6]. This algorithm is controlled by $l$. When $l \rightarrow \infty$, there is no denoising and the result is identical to the input signal. As $l = 0$ however, the total variation term plays an increasingly strong role, which forces the result to have smaller total variation, at the expense of being less like the initial noisy signal [7], [8] and [9].

*D. Bilateral filtering*

The bilateral filter is a non-linear approach which can reduce the noise of an image while preserving strong edges. This method can be traced back to 1995 with work of Aurich and Weule [10] in which the nonlinear Gaussian filter is improved. However, this method was later rediscovered by Tomasi and Manduchi, and Smith and Brady [11] who gave it its current name. This method can be formulated easily: each pixel is replaced by a weighted average of its neighbors. This filter is similar to the Gaussian filter. The difference is that the bilateral filter considers the difference in value with the neighbors to maintain edges while suppressing the noise. The main idea of the bilateral filter is that for a pixel to influence another one, it should not only have a similar value but also occupy a nearby location. The bilateral filter, denoted by BF[.], is defined by:

$$BF[I]_p = \frac{1}{W_p} \sum_{q \in S} G_{s_s}(\|p-q\|) G_{s_r}(|I_p - I_q|) \; I_q \quad (3)$$

where normalization factor is:

$$W_p = \sum_{q \in S} G_{s_s}(\|p-q\|) G_{s_r}(|I_p - I_q|) \quad (4)$$

and $G_{s_s}, G_{s_r}$ are the Gaussian functions with corresponding standard deviation that is defined by:

$$G_s(x) = \frac{1}{2\pi s^2} \exp(-\frac{x^2}{2s^2}) \quad (5)$$

The bilateral filter can be controlled by two different parameters, namely, $\sigma_s$ and $\sigma_r$. When the range parameter i.e., $\sigma_r$ increases, the bilateral filter gradually approximates Gaussian convolution more closely because the Gaussian function $G_{s_r}$ flattens and widens. In other words, it becomes nearly constant over the intensity value of the image. Increasing the spatial parameter i.e., $\sigma_s$ smoothes larger features of the image.

*E. Anisotropic Diffusion*

Anisotropic diffusion, also called Perona–Malik diffusion, is an approach aiming at suppressing the image noise without removing significant components of the image such as lines, edges, textures or other details which play vital role for the interpretation of the image [12]. When the isotropic diffusion is used, this method diffuses the pixel value in all of the directions, uniformly. This can lead to blurring of edges. On the contrary, for anisotropic diffusion, smoothing or diffusion is carried out depending on the direction of edges.

This method is similar to the TV method. The difference between these methods is that, in the anisotropic diffusion, the auxiliary function *c* is exploited to determine the amount of smoothing. In regions where the gradient ∇u is small, which are belong to noise or smooth regions, the diffusion is carried out strongly. On the contrary, in regions where the gradient ∇u is large, which are belong to edges and textures, the diffusion process is stopped [5] and [13]. The equation describing the method, is:

$$\frac{\partial I}{\partial t} = div(c(x,y,t)\nabla I) = \nabla c . \nabla I + c(x,y,t)\Delta I \quad (6)$$

where *c* is the auxiliary function. Perona and Malik proposed the following two auxiliary functions:

$$c(\|\nabla I\|) = e^{-(\|\nabla I\|/K)^2} \quad (7)$$

$$c(\|\nabla I\|) = \frac{1}{1+(\frac{\|\nabla I\|}{K})^2} \quad (8)$$

The constant *K* is usually chosen experimentally ( usually as a function of the noise of the image) to control the sensitivity to edges.

*F. Non-local mean algorithm*

The non-local means was proposed based on the assumption that the image contains self-similarity. This concept was developed for texture synthesis by Efros and Leung in 1999 [8]. Not only, neighboring pixels tend to have similar values, but also non-adjacent pixels will also have similar values.

Each pixel *p* of the noise free image obtained by the non-local means is calculated by the following formula [14]:

$$NL(V)(P) = \sum_{q \in V} w(p,q) V(q) \quad (9)$$

where *V* is the noisy image, and *w(p,q)* is the weight (0 ≤ *w(p,q)* ≤ 1). This weighting causes similar pixels have more influence than the non-similar ones. One of the function usually used for this purpose, is the Gaussian function. The level of filtering can be controlled by the standard deviation of the Gaussian function.

For removing periodic noise, filters are usually utilized in the frequency-domain. For this purpose, after finding the noise in the Fourier domain, the magnitude of those points is replaced with zero (using notch filters). In this way, the noise is eliminated. This method is not optimal, because a part of information from the main signal is removed during denoising. In this paper a method that solves this problem as much as possible, will be introduced. We will show that the proposed method achieves a better performance in comparison with the previous methods.

## II. METHODOLOGY

As it is mentioned in the previous section, there are several kinds of noise. Here, we discuss about a kind of noise having moire' pattern. A moiré pattern (/mwɑːrˈeɪ/; French: [mwaˈʁe]) or moiré fringes defined in mathematics and physics, is a secondary pattern formed when two transparent (usually identical) patterns on a curved or flat surface are displaced, or rotated a small amount from each other.

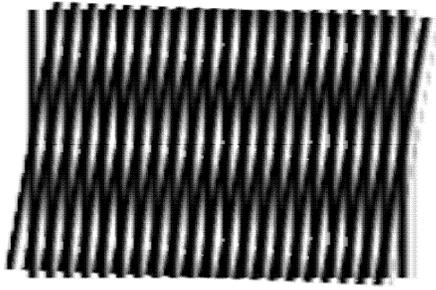

Fig. 1. Moire pattern

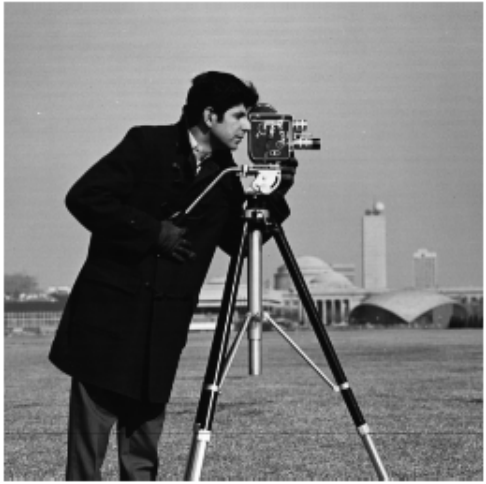

Fig. 2. First original test image (cameraman).

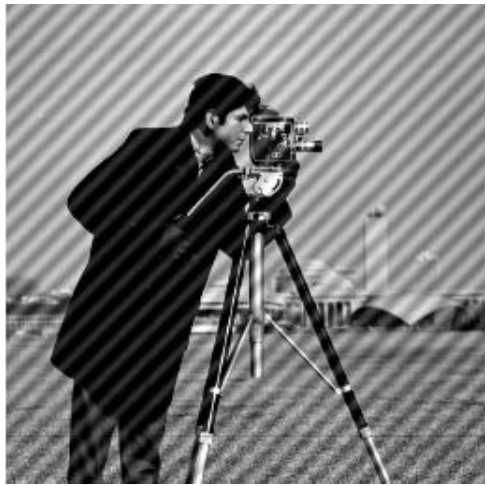

Fig. 3. First noisy image (noisy cameraman).

The suggested method in this paper is useful for the noise that has the moire' pattern. In Fig. (2) and Fig. (3) the original image and the noisy image which is corrupted by moire' pattern, are shown, respectively.

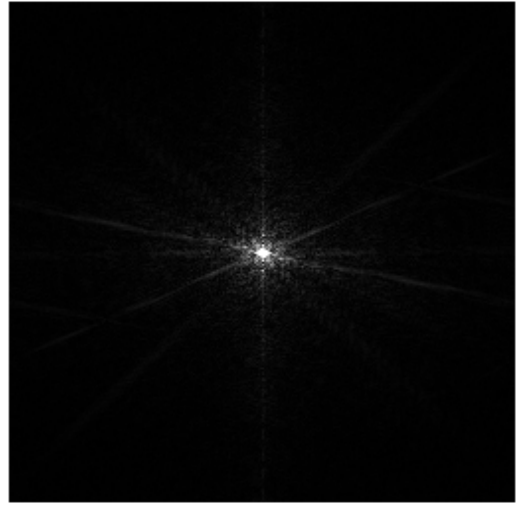

Fig. 4. Spectrum of the original cameraman

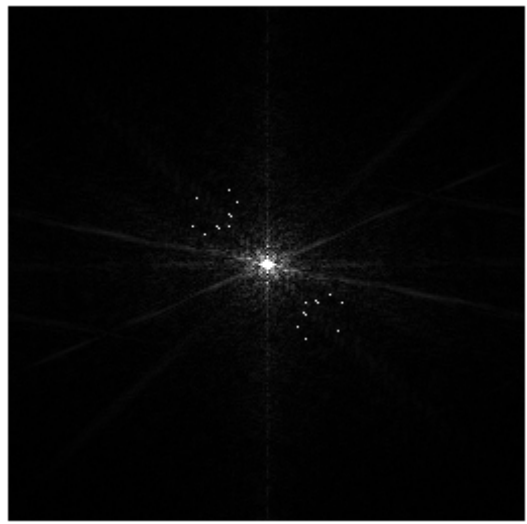

Fig. 5. Spectrum of the noisy image (noisy cameraman).

The amplitude of the Fourier transform (spectrum) of these two image are shown in Figs. (4) and (5), respectively. The moire' pattern creates some couples of impulses in the image's spectrum because of its sinusoidal nature, as it is clear in Fig. 5. In the conventional method some kinds of notch filters are used to eliminate the influence of noise in the spectrum of the noisy image [15]. Then, by using the inverse Fourier transform, the denoised image is obtained. But in that way, some information of the original image is lost during the denoising. In this paper we suggests a new method that uses some properties of spectrum of the natural signals.

First a 1D signal, i.e., a pulse is considered (see Fig. (6)). The amplitude of the Fourier transform of this signal is drawn in Fig. 7. It is obvious that the spectrum is continues.

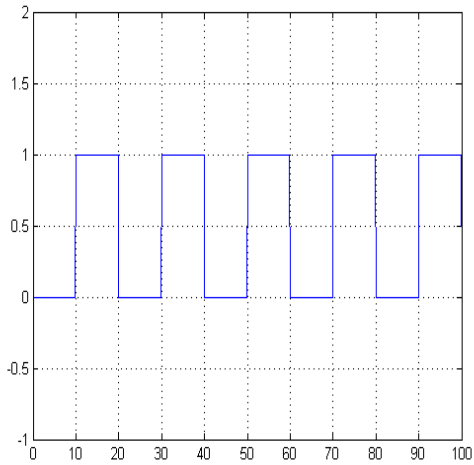

Fig. 6. 1D signal

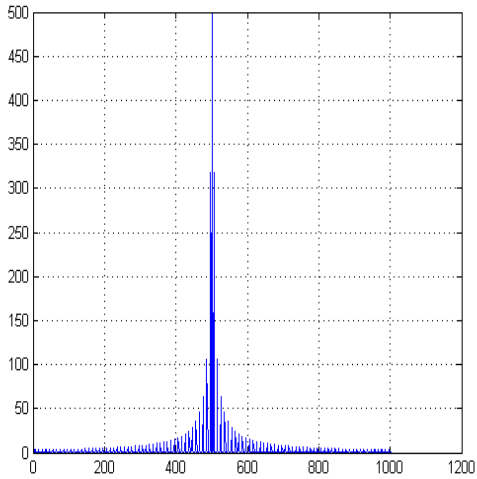

Fig. 7. Amplitute of the Fourior transform of signal in Fig. 1

In addition, the value of each point of the spectrum is similar to the value of surrounding points.

Now it can be generalized to 2D signals such as an image. In the proposed approach a median filter is used around the impulses to reduce the noise, instead of using notch filters. In Fig. 8 some of the "windows", related to the median filter, are shown. In this way, the spectrum of the image is estimated by using the value of neighbor points.

### III. RESULTS AND DISCUSSION

After using the median filter in that points, the inverse of Fourier transform is applied to the estimated spectrum. The denoised image for the first test image is shown in Fig. (9) which is similar to the original image (see Fig. 2).

The PSNR [16-19] value of the cameraman image in this technique is 56 dB, however in the conventional method the PSNR value is 51 dB. So our method has better performance

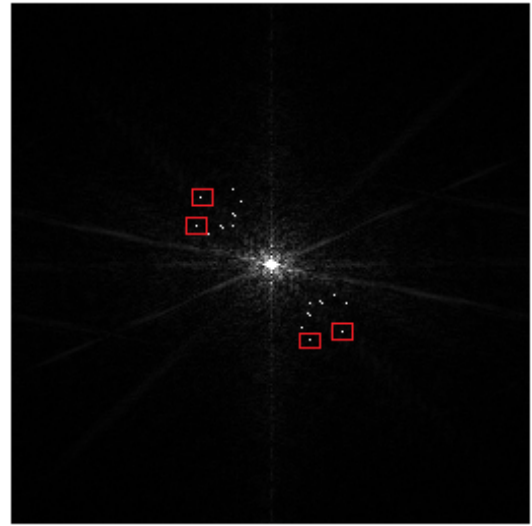

Fig .8. Windows of median filter

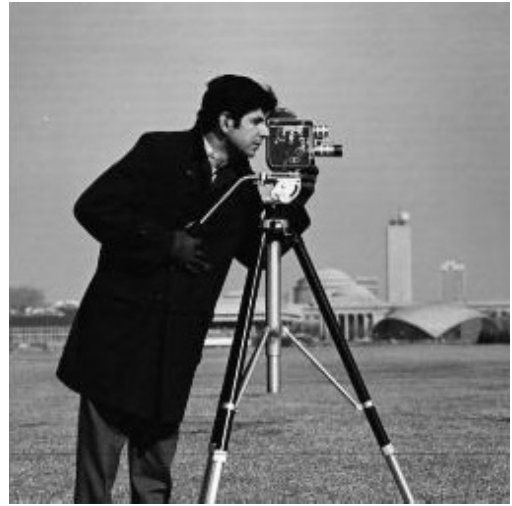

Fig. 9. The denoised image

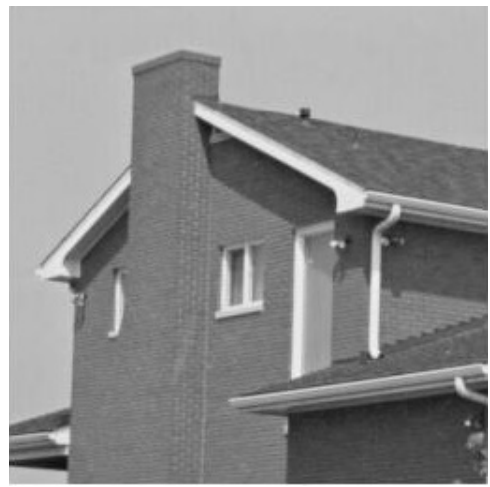

Fig. 10. Second test image (house).

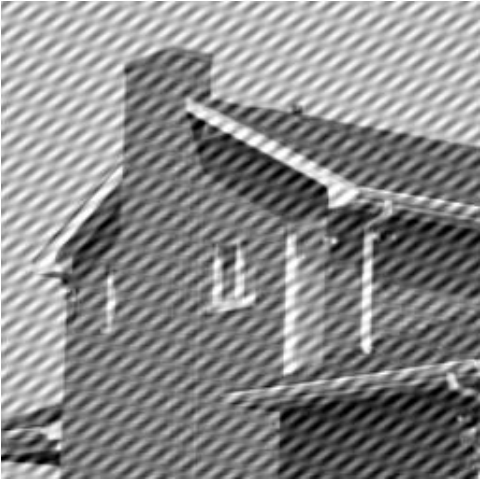

Fig. 11. Second test image with moire' pattern.

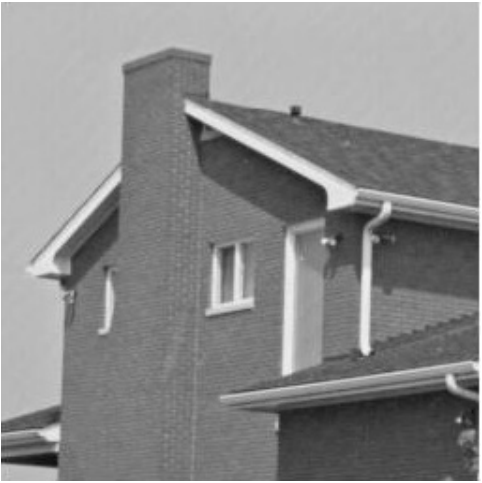

Fig. 12. Second test image after denoising

TABLE I. PSNR FOR SOME WELL KNOWN IMAGES.

| Images | PSNR in previous methods (dB) | PSNR in new method (dB) |
|---|---|---|
| cameraman | 50.18 | 55.99 |
| Barbara | 49.27 | 54.71 |
| Lena | 51.60 | 58.89 |
| Boat | 48.25 | 55.43 |
| House | 53.05 | 61.88 |
| Baboon | 52.64 | 60.07 |

in comparison with notch filters.

Figs. 10, 11 and 13 shows the second test image, noisy one and denoised image, respectively.

The result of the proposed method for some well-known image is shown in Table. I.

IV. CONCLUSION

In this paper a new technique for removing moire' pattern from images was proposed. Using median filter in this method gives a better performance in comparison with the conventional notch filters. Median filter eliminates noises while preserving most of information of images. We calculate PSNR for this method to show that this new method is more suitable for removing the moire' pattern.